\documentclass[conference]{IEEEtran}
\IEEEoverridecommandlockouts
\usepackage{algorithmic}
\usepackage{amsfonts}
\usepackage{amsmath,amssymb,amsfonts}
\usepackage{amsmath}
\usepackage{adjustbox}
\usepackage{cite}
\usepackage{epsfig}
\usepackage{graphicx}
\usepackage{textcomp}
\usepackage{url}
\usepackage{xcolor}

\title{The Case for RISP: \\ A Reduced Instruction Spiking Processor}
\author{James S. Plank, ChaoHui Zheng, Bryson Gullett, Nicholas Skuda, \\ Charles Rizzo, Catherine D. Schuman, Garrett S. Rose \\
{\em EECS Department, University of Tennessee, Knoxville, TN, USA}}
\date{}

\begin{document}

\maketitle

\begin{abstract}

In this paper, we introduce {\em RISP}, a reduced instruction spiking
processor.  While most spiking neuroprocessors are based on the
brain, or notions from the brain, we present the case for a spiking
processor that simplifies rather than complicates.  As such, it
features discrete integration cycles, configurable leak, and little else.  We
present the computing model of RISP and highlight the benefits of
its simplicity.  We demonstrate how it aids in
developing hand built neural networks for simple computational tasks,
detail how it may be employed to simplify neural networks built
with more complicated machine learning techniques, and demonstrate
how it performs similarly to other spiking neurprocessors.

\end{abstract}
\maketitle

\section{Introduction}

Since the 1960s, researchers have turned to the brain for inspiration
on how to build novel computing devices.  This has led to several
waves of research on spiking neuroprocessors~\cite{rjp:19:twb}.  In our
previous research, we have explored all facets of neuroprocessors, 
from fabrication and nanotechnology~\cite{boc:20:tsb,cbo:18:fcm}, 
to circuits and systems~\cite{fcm:21:amc,saa:20:dac} to applications
and simulations~\cite{dpd:18:fsd,kpm:21:bpn}, 
to data encoding and decoding~\cite{spb:19:nte},
to machine learning~\cite{pzh:20:rss,smp:20:eons}.
Informed especially by our development of several 
neuroprocessors~\cite{dsb:14:danna,mdb:18:dda,msp:20:cnd,cdr:16:hom},
we have been motivated to research what
happens when we simplify, rather than complicate.  The result is
{\em RISP}, a Reduced Instruction Spiking Processor.

RISP features integrate-and fire-neurons with discrete integration
cycles.  Synaptic delays are also discrete, and synaptic weights and
neuron thresholds may be configured as either discrete or analog.
Besides the aforementioned parameters, the only other configurable
feature of RISP is that neurons may either retain their activation
potentials indefinitely, or they may leak away completely at each
integration cycle.  There are no other complicating features of RISP
that are in many other neuroprocessors, such as plasticity,
refractory periods, custom integration techniques,
complicated leak models and learning rules.  In this paper we
define RISP and detail some of its attractive properties.  Our goal
is to demonstrate that a simple neuroprocessor like RISP is
viable alternative to more complicated neuroprocessors.

\section{Related Work}

There are several neuroprocessor models, typically implemented in simulation,
that are commonly used in the literature
to explore and evaluate spiking neural networks and 
neuromorphic-style computation.  These tend to be focused
on one particular domain at a time.  For example, there are a set of 
simulators that target computational neuroscience simulations, including
NEST~\cite{Gewaltig:NEST, eppler2009pynest}, 
NEURON~\cite{carnevale2006neuron}, Brian~\cite{goodman2008brian},
Brian2~\cite{stimberg2019brian}, 
Nengo~\cite{bekolay2014nengo}, GeNN~\cite{yavuz2016genn} and 
Brian2GeNN~\cite{stimberg2020brian2genn}.  Though these simulators
tend to have significant flexibility in the types of computation that can 
be achieved, with large variation in the types of neuron and synapse 
models that can be implemented, that flexibility often comes at the price 
of slow simulation times for even simple networks, making it difficult to 
use those simulators to evaluate new types of
algorithms~\cite{kulkarni2021benchmarking}. There are also simulators 
that primarily target machine learning-style computation with spiking 
neural networks, including BindsNET~\cite{Hazan2018} and 
NengoDL~\cite{rasmussen2019nengodl}. Because these simulation 
environments focus on the machine learning use case of spiking 
neural networks, their functionality is often restricted to accommodate 
that behavior.  Finally, there are simulation frameworks that primarily 
target particular neuromorphic hardware implementations, including 
NEMO~\cite{plagge2018nemo}, which targets IBM's 
TrueNorth~\cite{merolla2014million}, and NengoLoihi, which targets 
Intel's Loihi~\cite{davies2018loihi}. These simulators are focused on 
capturing the functionality of the underlying hardware and can also 
restrict the types of evaluations that can occur on those systems or 
perform prohibitively slowly~\cite{kulkarni2021benchmarking}.

\section{RISP Specification}

In RISP, each neuron has two configurable parameters: a
threshold, and whether or not the neuron leaks.  As with most spiking
neuroprocessors, each neuron stores an action potential, whose
value may be increased or decreased by spikes on incoming synapses.
Processing works in discrete timesteps, where incoming spikes are
integrated over the timestep, and it is the neuron's action
potential at the end of the timestep that determines whether or not
the neuron fires.  When a neuron is configured with leak enabled, then
if a neuron does not fire at the end of the integration cycle, its
action potential is reset to zero.  Without leak, neurons retain
their action potentials across time steps.

Because of the integration cycle of neurons, synapses only have
unit delays.  Like neurons, synapses may be configured to either have discrete
weights or analog weights.  There is no restriction on connectivity, 
or the number of synapses
that may come into or go out of a neuron.

Certain neurons may be designated as input neurons that may receive
spikes from the outside world.  Similarly, neurons may be specified
to be output neurons, whose spikes may be monitored by the outside world.

To be precise, a RISP network is defined by 8 sets:
\begin{enumerate}
\item $N$ Neurons:~$V = \{ v_0, v_1, ... v_{N-1}\}$
\item $N$ Thresholds:~$T = \{ t_0, t_1, ... t_{N-1}\} | t_i \in \cal{R} \; {\rm or} \; \cal{Z}$
\item $N$ Leaks:~$L = \{ l_0, l_1, ... l_{N-1}\} | t_i \in \{ T, F \}$
\item $M$ Synapses:~$E = \{ e_0, e_1, ... e_{M-1}\} | e_i = ( v_j \rightarrow v_k )$
\item $M$ Weights:~$W = \{ w_0, w_1, ... w_{M-1}\} | w_i \in \cal{R} \; {\rm or} \; \cal{Z}$
\item $M$ Delays:~$D = \{ d_0, d_1, ... d_{M-1}\} | d_i \in \cal{Z}$
\item Input neurons:~$I \subseteq V$.
\item Ouptut neurons:~$O \subseteq V$.
\end{enumerate}

\begin{figure}[ht]
\begin{center}
\includegraphics[scale=0.80]{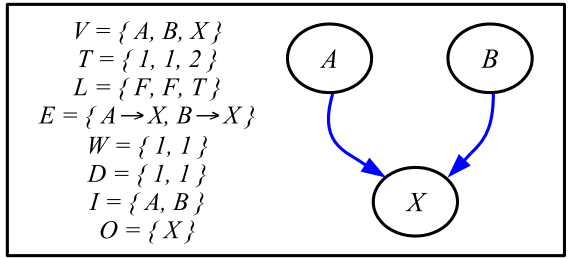}
\caption{\label{fig:example-and} An example RISP network that computes
the binary AND of inputs~$A$ and $B$.  The output neuron must be configured
with leak, so that when one input spikes and the other doesn't, the
action potential is reset to zero for the next problem.}
\end{center}
\end{figure}

Figure~\ref{fig:example-and} shows a very simple RISP network that computes the
binary AND of two inputs~$A$ and~$B$~\cite{pzs:21:snn}.  When spikes are applied
to both input neurons at timestep 0, then the output neuron $X$'s action potential is increased
to a value of two at timestep 1.  As such, it fires.
If a spike is applied to one of~$A$ or~$B$, but not to the other, then~$X$'s action potential
only reaches a value of one, and therefore it does not fire.  Moreover, since it is configured
with leak, its action potential is reset to zero, and
the network is ready to compute a new AND at the next timestep.

\section{Attractiveness of RISP}

RISP's simplicity is attractive in many pragmatic respects.  We detail
them in the sections below.

\subsection{Hand Constructing Networks}

Because of its simplicity, it is a straightforward matter to construct
RISP networks by hand, rather than by machine learning or other optimization
techniques.  This can be useful, either for solving simple problems, or for 
composing larger neural networks. 

\begin{figure}[ht]
\begin{center}
\includegraphics[scale=0.56]{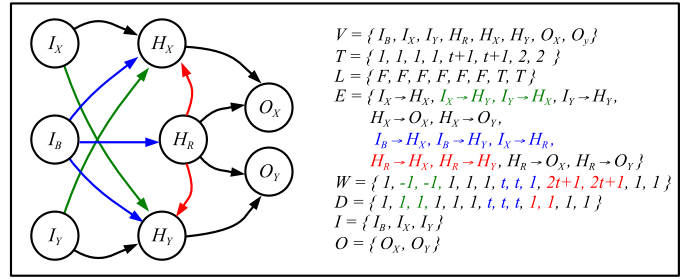}
\caption{\label{fig:max} RISP network that computes which input, $I_X$ or $I_Y$, fires more
          over an interval of~$t$ timesteps. The synapses are colored to facilitate reading
          the elements of the sets~$E$, $W$ and $D$.}
\end{center}
\end{figure}

We present an example in Figure~\ref{fig:max}.  This network calculates which input neuron,~$I_X$
or~$I_Y$, spikes more over an interval of~$t$ timesteps.
If~$I_X$ spikes more, then~$O_X$ spikes exactly once, at timestep~$t+1$.
If~$I_Y$ spikes more, then~$O_Y$ spikes exactly once, at timestep~$t+1$.
If they spike equally, then neither spikes, although it is a straightforward matter to 
add an extra synapse to break ties in favor of~$I_X$ or~$I_Y$.  There is a bias neuron
that is required to spike once at timestep 0.  The network clears itself, and may be
reused for more inputs at timestep~$t+1$.

This type of network is useful for composing spiking neural networks, perhaps created by
other methodologies (e.g.,~backpropagation~\cite{so:18:sl,svd:19:tdn} or genetic algorithms~\cite{smp:20:eons}).  For example, some neural networks, like the one we show later in section~\ref{sec:simp}, require their outputs to be intepreted by voting among
two output neurons.  The network in Figure~\ref{fig:max} performs this interpretation, which can in turn be used as inputs to other spiking neural networks, or even as control signals for a physical
device.

There have been other research projects that construct spiking neural networks that may be
implemented by RISP, either by targeting RISP explicitly, or by targeting a neural network that conforms to the RISP specifications.  For example, in our previous work, we have constructed 
RISP networks that perform binary operations on a variety of input/output encodings and decodings,
and RISP networks that convert between encodings/decodings~\cite{pzs:21:snn}.  The Whetstone
training methodology uses backpropagation and a ``sharpening'' activation function to train
deep spiking neural networks that may be implemented on RISP~\cite{svd:19:tdn}, and to reduce
the size of the resulting networks, we developed a convolutional ``kernel'' on RISP~\cite{pzh:20:rss}.  Monaco and Vidiola's integer factorization sieve~\cite{mv:17:if} may be implemented on RISP.

\subsection{Network Simplification}
\label{sec:simp}

\begin{figure}[ht]
\begin{center}
\includegraphics[scale=0.465]{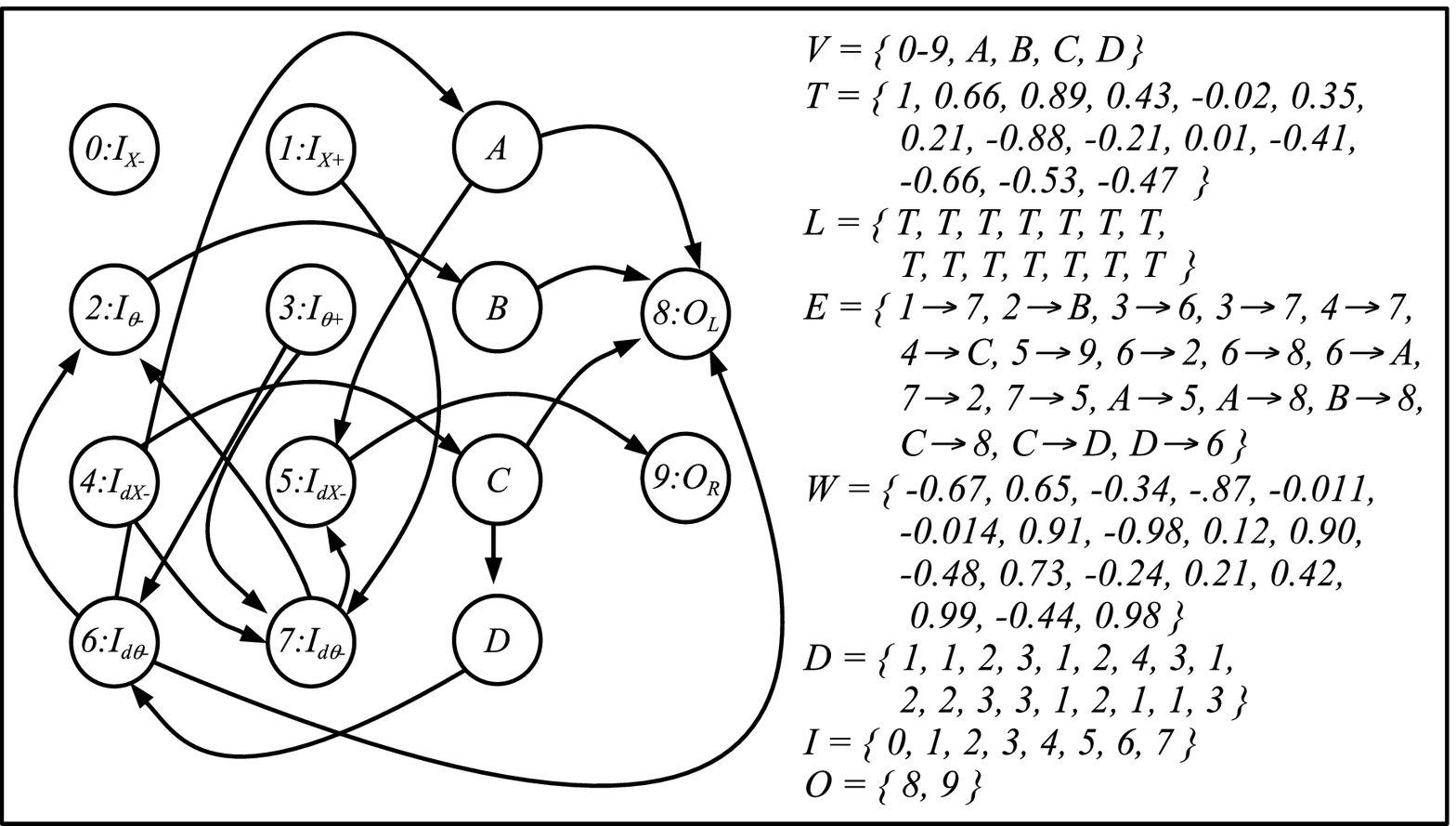}
\caption{\label{fig:pole-cluttered} RISP network to solve the cart-pole problem.  There
          are four inputs, $X$, $\theta$, $dX$, and $d\theta$, which report the cart's
          horizontal position, angle from vertical, horizontal speed and angle speed.
          Inputs are split over two neurons, and spike one to four times with unit values.
          There are two outputs, $O_L$ and~$O_R$ for boosting the cart left and right, and
          the decision is made by voting.  
          This network was created by the genetic algorithm, EONS.}
\end{center}
\end{figure}

RISP's simplicity makes it straightforward to analyze network behavior to simplify and
improve the networks.  As an example, consider the spiking neural network in 
Figure~\ref{fig:pole-cluttered}.  This network was trained by the genetic algorithm
EONS~\cite{smp:20:eons} to solve the cart-pole problem.  In this problem, there is
a cart enclosed on a fixed-size road, and on the cart is a pole on a hinge that can
fall left or right.  The goal of the problem is to push the cart left and right at fixed
power, and keep both the cart on the road and the pole balanced.

To solve this with RISP and EONS, we selected an input encoding technique where each input
value corresponds to two input neurons, with negative values spiked into one neuron (e.g.~$I_{X-}$)
and positive values spiked into another (e.g.~$I_{X+}$).  The magnitide of the value is
converted into a number of spikes, from one to four, and the spikes all have unit weights.

There are two output neurons,~$O_L$ and~$O_R$, which correspond to pushing the cart left and
right respectively.  The spiking neural network ``runs'' the application by working in discrete
simulated time intervals of 0.02 seconds.
At the beginning of each interval, the inputs are converted
to spikes and applied to the appropriate input neurons.  The neural network then runs for 50
timesteps, and the number of spikes on~$O_L$ and~$O_R$ are compared.  If~$O_R$ spikes more, then
the cart is pushed right.  Otherwise, the cart is pushed left.  (Interestingly, a network
like the one in Figure~\ref{fig:max} can be used to turn the votes into a single spike for pushing
the cart).

With EONS, we train the spiking neural network using ten random starting positions for the
cart and pole, and run the genetic algorithm until a network has been trained to keep the
cart on the road and the pole balanced for five simulated minutes in each of the ten starting
positions.  In a testing phase, the network in Figure~\ref{fig:pole-cluttered} keeps the
cart on the road and the pole balanced for an average of 4 minutes and 42 seconds on 100
randomly chosen starting positions.  As shown in the figure, the network has four hidden
neurons and 18 synapses.  The neuron thresholds and synapse weights are floating point numbers
whose values are rounded on the picture.

Because of the simplicity of RISP and of the input encoding, we can simplify this network with
the goals being:

\begin{itemize}
\item Reduce the number of neurons and synapses.
\item Set neuron thresholds and synapse weights to 1 or -1.
\end{itemize}

Both goals allow our networks to be implemented more simply, and to consume less power while
executing.  An example of this simplification is to consider the neurons $2$, $B$
and~$8$ in Figure~\ref{fig:pole-cluttered}.  Since the synapse~$2 \rightarrow B$ has a
weight of 0.65 and delay of 2, and neuron~$B$ has a threshold of -0.66, we may deduce that whenever neuron 2 fires, neuron~$B$ will fire one timestep later.  Further, because the synapse~$B \rightarrow 8$ is the only post-synapse to neuron~$B$, and its delay is two, we may simply delete neuron~$B$,
and replace its pre and post synapses with a synapse $2 \rightarrow 8$ whose delay is 3.

In a separate analysis, we observe that all four pre-synapses to neuron 8 have weights that exceed
the neuron's threshold of -0.21.  Therefore, we may set the weights of all of these synapses
equal to one, and set the threshold of neuron 8 to one as well.

\begin{figure}[ht]
\begin{center}
\includegraphics[scale=0.5]{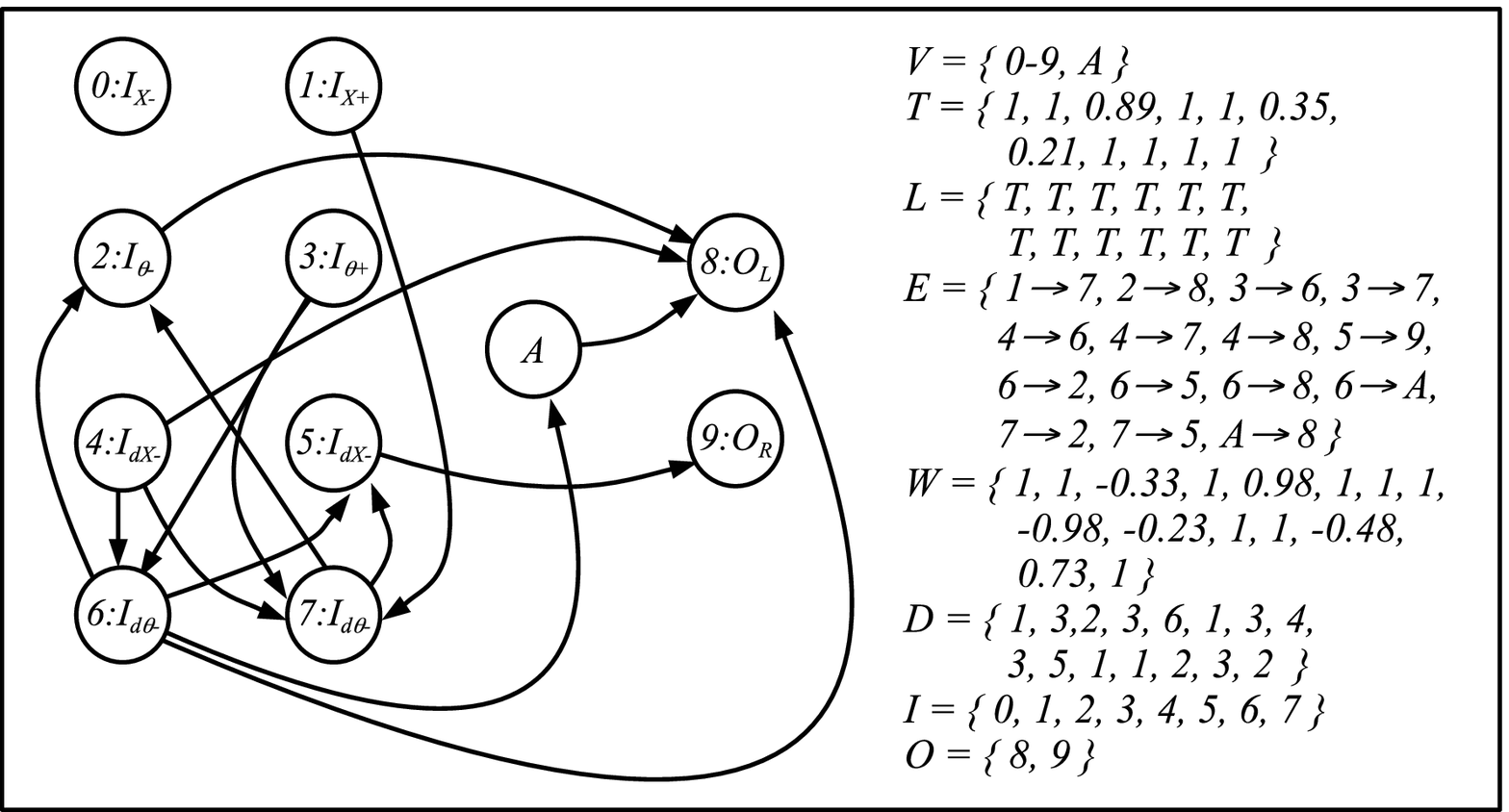}
\caption{\label{fig:pole-clean} RISP network that is equivalent to the network
          in Figure~\ref{fig:pole-cluttered}, but has been simplified.  It has three
          fewer neurons and three fewer synapses.  Moreover, seven of the neuron thresholds
          and nine of the synapse weights have been set to one.  Therefore, this network
          is simpler to implement and should require less power consumption.  Its performance
          on the application is the same.}
\end{center}
\end{figure}

By applying observations such as these, we may simplify the 
network in Figure~\ref{fig:pole-cluttered} into the one in Figure~\ref{fig:pole-clean}.
This network has only one hidden neuron, a reduction of 75 percent, and 15 synapses,
a reduction of 17 percent.  Moreover, seven of the neuron thresholds (64 percent) and
nine of the synapse weights (60 percent) have been set to one.

One of our current research projects is to implement these and similar observations to
simplify RISP networks.  We are also attempting stochastic and enumeration techniques
to simplify the networks using empirical observations.  These techniques are enabled
by the fact that RISP is simple, and does not have features such as plasticity, complex
learning rules and refractory periods, that complicate the analysis of the networks.

\subsection{Genetic Training and Simulation Speed}

Over the past seven years, the TENNLab research group has performed many explorations
of real-time control applications with spiking neuroprocessors~\cite{prs:19:tsl,afd:20:ggr,mbd:17:neon,spb:19:nte}.  One of the easiest of these is the cart-pole application mentioned in
section~\ref{sec:simp}.  In this section, we perform a small experiment using a genetic
optimization of spiking neural networks on the cart-pole application,
to see whether RISP trains similarly to other spiking processors.  We use the same input
enocoding as in section~\ref{sec:simp}.
We perform six sets of
optimizations using the EONS genetic algorithm and the TENNLab neuromorphic computing framework~\cite{smp:20:eons,psb:18:ten,spp:21:sfc}.  These are listed in Table~\ref{tab:proc}.

\begin{table}[ht]
\begin{center}
\begin{tabular}{c|cccc}
Label & Processor & Values & Leak & Features \\
\hline
RISP-A & RISP & Analog & No & None \\
RISP-D & RISP & Discrete & No & None \\
RISP-A-L & RISP & Analog & Yes & None \\
RISP-D-L & RISP & Discrete & Yes & None \\
Caspian & Caspian & Discrete & Yes & None \\
DANNA2 & DANNA2 & Discrete & Yes & Refractory \\
\multicolumn{5}{c}{} \\
\end{tabular}
\caption{\label{tab:proc} Neuroprocessors tested in a genetic training
experiment on the cart-pole problem.}
\end{center}
\end{table}

The first four processors listed are the four variants of RISP.  The fourth is Caspian~\cite{msp:20:cnd}, developed at Oak Ridge National Laboratory and implemented on FPGA's, with a very fast
simulator.  The fifth is DANNA2, developed by TENNLab and implemented on FPGA's, with a
VLSI workflow~\cite{mdb:18:dda}.  

We performed a simple genetic optimization composed of populations of 500 networks optimizing
over 100 epochs, with 10 random starting positions of the cart and pole.  For each neuroprocessor
in Table~\ref{tab:proc}, we performed 100 independent optimizations.

\begin{figure}[ht]
\begin{center}
\begin{tabular}{cc}
\includegraphics[scale=1.00]{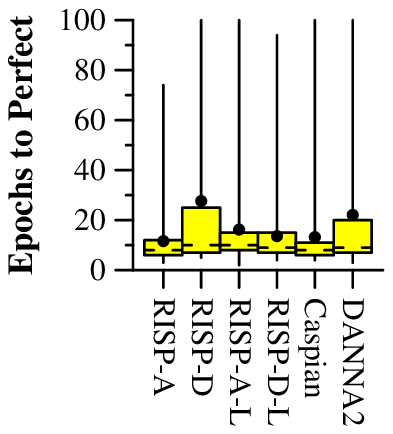} &
\includegraphics[scale=1.00]{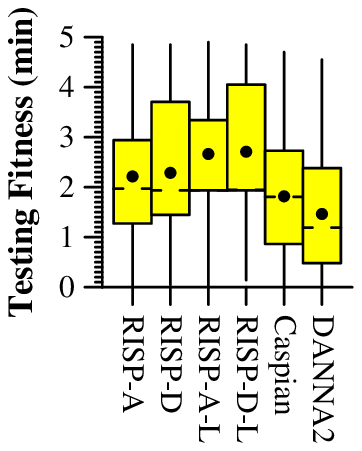} \\
\end{tabular}
\caption{\label{fig:eptest} Results of training and testing the cart-pole application
using the EONS genetic algorithm.}
\end{center}
\end{figure}

In Figure~\ref{fig:eptest}, we show both training and testing results for the neuroprocessors.

Because cart-pole is such a simple application, all of the neuroprocessors train to a perfect
fitness in nearly every run.  The left graph in Figure~\ref{fig:eptest} shows a tukey plot
of the number of epochs until perfect training fitness is achieved (keeping the pole balanced
and the cart in position for 5 simulated minutes for each of the 10 training episodes).
RISP's training success is similar to the others, with RISP-D requiring more epochs than the
others.  

To test, we run the final trained network on 100 episodes of cart-pole with different starting
positions than the training runs, and record their average values.  A ``perfect'' training run
lasts five simulated minutes.  The values for the 100 episodes are averaged, and the 
tukey plots for the 100 runs are shown on the right side of Figure~\ref{fig:eptest}.  RISP's
testing performance was superior to the others, on average, but we are hesitant
to draw conclusions without performing a larger study.

The speed of simulation impacts the speed of training.  Caspian, and the three RISP processors with the exception of RISP-D took between 40 and 83 minutes to complete all 100 of their training
runs.  RISP-D was slower, at nearly 4 hours, most likely because of the increased number of training
epochs.  DANNA2's simulator is much slower, requiring 27 hours to complete.  For reference, 
our NengoLoihi runs took a day to complete an average of 5 epochs, while dedicating 10 processors
to each training run (the others used a single processor).

We conclude from this study that RISP is viable alternative to the other neuroprocessors
for training via genetic algorithm.  Its simple features do not hinder the success of training,
and its simulation speed is excellent.

\section{Conclusion}

In this paper, we introduce RISP, a Reduced Instruction Spiking Processor, whose goal is
to explore the effectiveness of simplifying spiking neuroprocessors.  RISP features 
integrate-and-fire neurons, unit integration cycles and therefore unit synaptic delays,
and either analog or discrete weights and thresholds.  The only other feature is that
neurons may leak their action potentials at each cycle, or they may retain them.

In this paper, we demonstrate how RISP's simplicity enables the hand-construction of networks,
and how it allows us to reason about pre-trained networks to simplify them.  Finally, we
perform a very simple experiment with genetic training on a control application to show that
RISP trains on par with other neuroprocessors that contain more features.

Our near-future plan with RISP is to implement a network compiler that renders RISP networks
on inexpensive microcontrollers like the RP2040 (\$5) and FPGA's like the UPduino (\$25).
We anticipate that RISP's simplicity should lead to very clean implementations, featuring
high density of neurons and synapses on these inexpensive devices.  The devices
may then be used in physical experiments such as the ones we performed with 
the robots NeoN and GRANT~\cite{mbd:17:neon,afd:20:ggr}.

The bottom line is that simplicity in a spiking neuroprocessor has multiple advantages,
and is worth exploring as an alternative to neuroprocessors with more complicated, albeit
potentially bio-realistic features.  Our goal with RISP is for it to be the ``simple'' option
as a spiking neuroprocessor in applications and scenarios that desire to explore spiking
neuroprocessors as their solutions.

\section{Acknowledgments}

This research was supported in part by an Air
Force Research Laboraty Information Directorate grant (FA8750-19-1-0025).

\bibliographystyle{plain}
\bibliography{neuro}

\begin{thebibliography}{10}

\bibitem{afd:20:ggr}
J.~D. Ambrose, A.~Z. Foshie, M.~E. Dean, J.~S. Plank, G.~S. Rose, J.~P.
  Mitchell, C.~D. Schuman, and G.~Bruer.
\newblock {GRANT}: Ground roaming autonomous neuromorphic targeter.
\newblock In {\em IJCNN: The International Joint Conference on Neural
  Networks}, July 2020.

\bibitem{boc:20:tsb}
K.~Beckmann, W.~Olin-Ammentorp, G.~Chakma, S.~Amer, G.~S. Rose, C.~Hobbs,
  J.~{Van Nostrand}, M.~Rodgers, and N.~C. Cady.
\newblock Towards synaptic behavior of nanoscale reram devices for neuromorphic
  computing applications.
\newblock {\em ACM Journal on Emerging Technologies in Computing Systems},
  16(2), April 2020.

\bibitem{bekolay2014nengo}
Trevor Bekolay, James Bergstra, Eric Hunsberger, Travis DeWolf, Terrence~C
  Stewart, Daniel Rasmussen, Xuan Choo, Aaron Voelker, and Chris Eliasmith.
\newblock Nengo: a python tool for building large-scale functional brain
  models.
\newblock {\em Frontiers in neuroinformatics}, 7:48, 2014.

\bibitem{cbo:18:fcm}
N.~C. Cady, K.~Beckmann, W.~Olin-Ammentorp, G.~Chakma, S.~Amer, R.~Weiss,
  S.~Sayyaparaju, M.~Adnan, J.~Murray, M.~Dean, J.~Plank, G.~Rose, and J.~{Van
  Nostrand}.
\newblock Full {CMOS}-memristor implementation of a dynamic neuromorphic
  architecture.
\newblock In {\em 43rd Annual GOMACTech Conference}, Miami, March 2018.

\bibitem{carnevale2006neuron}
Nicholas~T Carnevale and Michael~L Hines.
\newblock {\em The {NEURON} book}.
\newblock Cambridge University Press, 2006.

\bibitem{cdr:16:hom}
G.~Chakma, Mark~E. Dean, G.~S. Rose, K.~Beckmann, H.~Manem, and N.~Cady.
\newblock A hafnium-oxide memristive dynamic adaptive neural network array.
\newblock In {\em International Workshop on Post-Moore's Era Supercomputing
  (PMES)}, Salt Lake City, UT, November 2016.

\bibitem{davies2018loihi}
Mike Davies, Narayan Srinivasa, Tsung-Han Lin, Gautham Chinya, Yongqiang Cao,
  Sri~Harsha Choday, Georgios Dimou, Prasad Joshi, Nabil Imam, Shweta Jain,
  et~al.
\newblock Loihi: A neuromorphic manycore processor with on-chip learning.
\newblock {\em IEEE Micro}, 38(1):82--99, 2018.

\bibitem{dsb:14:danna}
M.~E. Dean, C.~D. Schuman, and J.~D. Birdwell.
\newblock Dynamic adaptive neural network array.
\newblock In {\em 13th International Conference on Unconventional Computation
  and Natural Computation (UCNC)}, pages 129--141, London, ON, July 2014.
  Springer.

\bibitem{dpd:18:fsd}
A.~W. Disney, J.~S. Plank, and M.~Dean.
\newblock Four simulators of the {DANNA} neuromorphic computing architecture.
\newblock In {\em International Conference on Neuromorphic Computing Systems},
  Knoxville, TN, July 2018. ACM.

\bibitem{eppler2009pynest}
Jochen~M Eppler, Moritz Helias, Eilif Muller, Markus Diesmann, and Marc-Oliver
  Gewaltig.
\newblock Py{NEST}: a convenient interface to the {NEST} simulator.
\newblock {\em Frontiers in neuroinformatics}, 2:12, 2009.

\bibitem{fcm:21:amc}
A.~Z. Foshie, N.~N. Chakraborty, J.~J. Murray, T.~J. Fowler, M.~S.~A. Shawkat,
  and G.~S. Rose.
\newblock A multi-context neural core design for reconfigurable neuromorphic
  arrays.
\newblock In {\em IEEE Computer Society Annual Symposium on VLSI (ISVLSI)},
  pages 67--72, July 2021.

\bibitem{Gewaltig:NEST}
Marc-Oliver Gewaltig and Markus Diesmann.
\newblock {NEST} ({NE}ural {S}imulation {T}ool).
\newblock {\em Scholarpedia}, 2(4):1430, 2007.

\bibitem{goodman2008brian}
Dan~FM Goodman and Romain Brette.
\newblock Brian: a simulator for spiking neural networks in python.
\newblock {\em Frontiers in neuroinformatics}, 2:5, 2008.

\bibitem{Hazan2018}
Hananel Hazan, Daniel~J. Saunders, Hassaan Khan, Devdhar Patel, Darpan~T.
  Sanghavi, Hava~T. Siegelmann, and Robert Kozma.
\newblock Binds{NET}: A machine learning-oriented spiking neural networks
  library in python.
\newblock {\em Frontiers in Neuroinformatics}, 12:89, 2018.

\bibitem{kpm:21:bpn}
S.~Kulkarni, M.~Parsa, J.~P. Mitchell, and C.~D. Schuman.
\newblock Benchmarking the performance of neuromorphic and spiking neural
  simulators.
\newblock {\em Neurocomputing}, 447:145--160, 2021.

\bibitem{kulkarni2021benchmarking}
Shruti~R Kulkarni, Maryam Parsa, J~Parker Mitchell, and Catherine~D Schuman.
\newblock Benchmarking the performance of neuromorphic and spiking neural
  network simulators.
\newblock {\em Neurocomputing}, 447:145--160, 2021.

\bibitem{merolla2014million}
Paul~A Merolla, John~V Arthur, Rodrigo Alvarez-Icaza, Andrew~S Cassidy, Jun
  Sawada, Filipp Akopyan, Bryan~L Jackson, Nabil Imam, Chen Guo, Yutaka
  Nakamura, et~al.
\newblock A million spiking-neuron integrated circuit with a scalable
  communication network and interface.
\newblock {\em Science}, 345(6197):668--673, 2014.

\bibitem{mbd:17:neon}
J.~P. Mitchell, G.~Bruer, M.~E. Dean, J.~S. Plank, G.~S. Rose, and C.~D.
  Schuman.
\newblock {NeoN}: Neuromorphic control for autonomous robotic navigation.
\newblock In {\em IEEE 5th International Symposium on Robotics and Intelligent
  Sensors}, pages 136--142, Ottawa, Canada, October 2017.

\bibitem{mdb:18:dda}
J.~P. Mitchell, M.~E. Dean, G.~Bruer, J.~S. Plank, and G.~S. Rose.
\newblock {DANNA 2}: Dynamic adaptive neural network arrays.
\newblock In {\em International Conference on Neuromorphic Computing Systems},
  Knoxville, TN, July 2018. ACM.

\bibitem{msp:20:cnd}
J.~P. Mitchell, C.~D. Schuman, R.~M. Patton, and T.~E. Potok.
\newblock Caspian: A neuromorphic development platform.
\newblock In {\em NICE: Neuro-Inspired Computational Elements Workshop}. ACM,
  March 2020.

\bibitem{mv:17:if}
J.~V. Monaco and M.~M. Vindiola.
\newblock Integer factorization with a neuromorphic sieve.
\newblock {\em CoRR}, abs/1703.03768, 2017.

\bibitem{plagge2018nemo}
Mark Plagge, Christopher~D Carothers, Elsa Gonsiorowski, and Neil Mcglohon.
\newblock Nemo: A massively parallel discrete-event simulation model for
  neuromorphic architectures.
\newblock {\em ACM Transactions on Modeling and Computer Simulation (TOMACS)},
  28(4):1--25, 2018.

\bibitem{prs:19:tsl}
J.~S. Plank, C.~Rizzo, K.~Shahat, G.~Bruer, T.~Dixon, M.~Goin, G.~Zhao,
  J.~Anantharaj, C.~D. Schuman, M.~E. Dean, G.~S. Rose, N.~C. Cady, and J.~{Van
  Nostrand}.
\newblock The {TENNLab} suite of {LIDAR}-based control applications for
  recurrent, spiking, neuromorphic systems.
\newblock In {\em 44th Annual GOMACTech Conference}, Albuquerque, March 2019.

\bibitem{psb:18:ten}
J.~S. Plank, C.~D. Schuman, G.~Bruer, M.~E. Dean, and G.~S. Rose.
\newblock The {TENNLab} exploratory neuromorphic computing framework.
\newblock {\em IEEE Letters of the Computer Society}, 1(2):17--20, July-Dec
  2018.

\bibitem{pzh:20:rss}
J.~S. Plank, J.~Zhao, and B.~Hurst.
\newblock Reducing the size of spiking convolutional neural networks by trading
  time for space.
\newblock In {\em IEEE International Conference on Rebooting Computing (ICRC)},
  pages 116--125. IEEE, December 2020.

\bibitem{pzs:21:snn}
J.~S. Plank, C.~Zheng, C.~D. Schumann, and C.~Dean.
\newblock Spiking neuromorphic networks for binary tasks.
\newblock In {\em International Conference on Neuromorphic Computing Systems
  (ICONS)}, pages 1--8. ACM, 2021.

\bibitem{rasmussen2019nengodl}
Daniel Rasmussen.
\newblock Nengo{DL}: Combining deep learning and neuromorphic modelling
  methods.
\newblock {\em Neuroinformatics}, 17(4):611--628, 2019.

\bibitem{rjp:19:twb}
K.~Roy, A.~Jaiswal, and P.~Panda.
\newblock Towards spike-based machine intelligence with neuromorphic computing.
\newblock {\em Nature}, 575:607 -- 617, 2019.

\bibitem{saa:20:dac}
S.~Sayyaparaju, M.~M. Adnan, S.~Amer, and G.~S. Rose.
\newblock Device-aware circuit design for robust memristive neuromorphic
  systems with stdp-based learning.
\newblock {\em ACM Journal on Emerging Technologies in Computing Systems},
  16(3), May 2020.

\bibitem{smp:20:eons}
C.~D. Schuman, J.~P. Mitchell, R.~M. Patton, T.~E. Potok, and J.~S. Plank.
\newblock Evolutionary optimization for neuromorphic systems.
\newblock In {\em NICE: Neuro-Inspired Computational Elements Workshop}, 2020.

\bibitem{spb:19:nte}
C.~D. Schuman, J.~S. Plank, G.~Bruer, and J.~Anantharaj.
\newblock Non-traditional input encoding schemes for spiking neuromorphic
  systems.
\newblock In {\em IJCNN: The International Joint Conference on Neural
  Networks}, pages 1--10, Budapest, 2019.

\bibitem{spp:21:sfc}
C.~D. Schuman, J.~S. Plank, M.~Parsa, S.~R. Kulkarni, N.~Skuda, and J.~P.
  Mitchell.
\newblock A software framework for comparing training approaches for spiking
  neuromorphic systems.
\newblock In {\em IJCNN: The International Joint Conference on Neural
  Networks}, pages 1--10, July 2021.

\bibitem{svd:19:tdn}
W.~Severa, C.~M. Vineyard, R.~Dellana, S.~J. Verzi, and J.~B. Aimone.
\newblock Training deep neural networks for binary communication with the
  {W}hetstone method.
\newblock {\em Nature Machine Intelligence}, 1:86--94, January 2019.

\bibitem{so:18:sl}
S.~B. Shrestha and G.~Orchard.
\newblock {SLAYER}: Spike layer error reassignment in time.
\newblock In S.~Bengio, H.~Wallach, H.~Larochelle, K.~Grauman, N.~Cesa-Bianchi,
  and R.~Garnett, editors, {\em Advances in Neural Information Processing
  Systems 31}, pages 1412--1421. Curran Associates, Inc., 2018.

\bibitem{stimberg2019brian}
Marcel Stimberg, Romain Brette, and Dan~FM Goodman.
\newblock Brian 2, an intuitive and efficient neural simulator.
\newblock {\em Elife}, 8:e47314, 2019.

\bibitem{stimberg2020brian2genn}
Marcel Stimberg, Dan~FM Goodman, and Thomas Nowotny.
\newblock {B}rian2{G}e{NN}: accelerating spiking neural network simulations
  with graphics hardware.
\newblock {\em Scientific Reports}, 10(1):1--12, 2020.

\bibitem{yavuz2016genn}
Esin Yavuz, James Turner, and Thomas Nowotny.
\newblock Ge{NN}: a code generation framework for accelerated brain
  simulations.
\newblock {\em Scientific reports}, 6:18854, 2016.

\end{thebibliography}

\end{document}